# Un modèle pour la représentation des connaissances temporelles dans les documents historiques

Sahar Aljalbout, Gilles Falquet

Centre Universitaire d'informatique, Université de Genève
`saharjalbout@gmail.com`
`Gilles.falquet@unige.ch`

**Résumé** : Traiter et publier les données des sciences historiques dans le web sémantique constitue un défi intéressant où la représentation des aspects temporels joue un rôle clé. Nous proposons dans cet article un modèle de représentation des connaissances temporelles adapté au travail sur les documents historiques. Ce modèle est basé sur la notion de fluent que l'on représente dans des graphes RDF. Nous montrons comment ce modèle permet de représenter les connaissances nécessaires aux historiens et de raisonner sur celles-ci à l'aide des langages SWRL et SPARQL. Ce modèle est en cours d'utilisation dans un projet de numérisation, d'étude et de publication des manuscrits du linguiste Ferdinand de Saussure.

**Mots-clés** : web sémantique, représentation des connaissances temporelles, raisonnement temporel, documents historiques, ontologies historiques

## 1 Introduction

La représentation des connaissances temporelles avec les langages du web sémantique reste encore un défi, en particulier en terme de simplicité d'utilisation et de capacité de raisonnement. Pourtant, les sciences historiques et les données qu'elles produisent constituent un vaste domaine d'application dans lequel de nouvelles techniques de représentation, de raisonnement et de publication sur le web sémantique restent à créer.

Dans cet article nous présentons un modèle pour la représentation et le raisonnement temporel dans l'analyse des documents historiques. Ce modèle est actuellement appliqué dans le cadre d'un projet interdisciplinaire de publication savante des manuscrits de Ferdinand de Saussure. Saussure (1857-1913) est un linguiste suisse connu comme l'un des fondateurs de la linguistique moderne. De son vivant, Saussure n'a que très peu publié ses travaux. Il a par contre laissé de nombreux textes manuscrits (représentant plus de 30 000 pages) qui constituent une ressource d'une très grande richesse dans le domaine de la linguistique et de son histoire. Cependant leur étude est rendue complexe par le fait que nombre de ces textes ne sont pas datés et que la terminologie utilisée (et créée) par Saussure change considérablement au cours du temps. D'où l'intérêt de disposer d'un modèle de connaissances et de techniques d'inférence temporelles pour aider les experts saussuriens à indexer sémantiquement ces textes, à les dater ou à reconstituer leur séquence temporelle et en fin de compte à les comprendre.

En outre, la communauté des humanités numériques porte un grand intérêt à la publication des connaissances sur le web sémantique. C'est pourquoi nous avons décidé d'utiliser les technologies du web sémantique comme cadre d'implémentation du modèle.

## 2  Les besoins pour une représentation adéquate des connaissances historiques

Nous avons effectué une analyse détaillée des besoins des historiens de la linguistique concernant la dimension temporelle d'une base de connaissance. Le résultat de cette analyse a permis de classer les besoins de modélisation temporelle liée aux documents historiques selon quatre axes:

*Représentation des entités contextuelles présentes dans les documents.* La détermination des personnages, lieux, évènements auxquels les documents font référence permet d'exploiter le contexte historique pour mieux comprendre les documents.

*Représentation des changements dans le contexte historique au cours du temps.* Des propriétés telles que le lieu de résidence, la fonction ou les liens de collaboration d'une personne évoluent au cours du temps.

*Représentation de la cause des changements (actions, évènements, ...)*

*Représentation des différentes terminologies utilisées dans les documents.* La compréhension du contenu d'un document historique est toujours relative. Elle dépend de la terminologie employée par l'auteur et de la conscience qu'a le lecteur de l'usage de cette terminologie.

La prise en comte de ces besoins a guidé la construction et la validation de la partie structurelle de notre modèle.

## 3  Etat de l'art

La prise en compte du temps dans les bases de connaissance, et en particulier dans le Web sémantique a donné lieu à de nombreux travaux, dont nous ne mentionnerions que les plus pertinents pour notre étude.

### 3.1  Développements théoriques

Dans (Gutierrez et al., 2007) puis (Motik, 2012) les auteurs traitent du temps de validité en RDF et OWL. Les graphes temporels qu'ils proposent contiennent des étiquettes de validité temporelle associées à chaque triplet. Ils définissent une notion de conséquence logique entre ces graphes et des opérations d'interrogation
Les travaux sur la logique de description temporelle (Lutz et Wolter, 2008) introduisent quand à eux des opérateurs qui permettent de définir des concepts intrinsèquement temporels (p.ex. le concept d'*être mortel*) ..
Il existe également des approches tel *Event Calculus* (Mueller, 2008) qui ont pour but le raisonnement sur les actions et les changements qu'elles provoquent.
Un autre axe de recherche consiste à utiliser la notion de version d'ontologies (Klein et Fensel, 2001) pour représenter l'évolution de la connaissance (et de la terminologie).

### 3.2  Implémentation dans les langages existants

D'un point de vue plus pratique, certaines techniques ont été proposées pour utiliser les langages et systèmes existants pour la modélisation temporelle. Elles essaient en général de surmonter les limitations imposées par le seul usage des prédicats binaires en RDF et OWL. Parmi celles-ci on trouve :
- les *named graphs* (Tappolet et Bernstein, 2009) où chaque graphe contient les triplets qui sont vrais pour un intervalle de temps spécifié

- l'utilisation de patrons pour la représentation de relations n-aires en OWL et RDF(S) (Noy et Rector, 2006). La dimension temporelle peut alors être ajoutée à chaque relation binaire.
- les 4D-fluents (Welty et Fikes, 2006) où chaque concept temporel est représenté comme un objet à 4 "dimensions" avec comme quatrième dimension le temps.

### 3.3 Discussion

Les travaux présentés précédemment sont de natures différentes, riches et variés mais on constate que les travaux théoriques n'ont pas abouti à des langages et systèmes largement acceptés ou utilisés. Les solutions pragmatiques souffrent quand à elles de divers défauts allant de la prolifération d'objets dans les graphes à l'absence de mécanismes de raisonnement. Elles ne prennent pas non plus en considération tous les besoins de modélisation historique définis dans la section 2..

## 4 Un modèle temporel pour la représentation des connaissances historiques

Le modèle que nous avons défini pour représenter les connaissances liées aux manuscrits historiques comprend la représentation des manuscrits eux-mêmes (images, transcriptions, annotations, etc.), la représentation du contexte historique (personnes, lieux, évènements, terminologies employées, etc.) et les liens de référence entre manuscrits et entités du contexte.

### 4.1 Le modèle temporel et sa réalisation dans le Web sémantique

Pour représenter les changements dans le contexte historique et leurs causes nous proposons d'utiliser la notion de fluent (McCarthy et Hayes, 1969), (Mueller, 2008). Un fluent (propositionnel) est défini comme une fonction qui associe une valeur de vérité à un énoncé pour un instant donné.

Nous appellerons *assertion de fluent* un énoncé qui indique qu'un fluent est vrai pendant un intervalle de temps donné. Par exemple, l'énoncé « *Saussure a vécu en Allemagne entre 1876 et 1881* », indique que le fluent « *Saussure vit en Allemagne* » est vrai pendant toute la période 1876-1881. Par contre un énoncé tel que « *Saussure est né en 1857* », bien que temporel, ne concerne pas un fluent car la relation *est né en* entre Saussure et 1857 n'est pas susceptible de changer au cours du temps.

Le problème de la représentation des fluents dans le Web sémantique est lié aux langages de représentation tels que RDF qui ne supportent que les relations binaires. Même si les relations qui varient au cours du temps sont binaires, la représentation de l'intervalle temporel durant lequel la relation est vraie nécessite un troisième argument ou une réification. D'autre part, il est également nécessaire de pouvoir représenter la cause qui a initialement rendu vrai le fluent et celle qui l'a rendu faux à la fin de l'intervalle (quand elles sont connues).

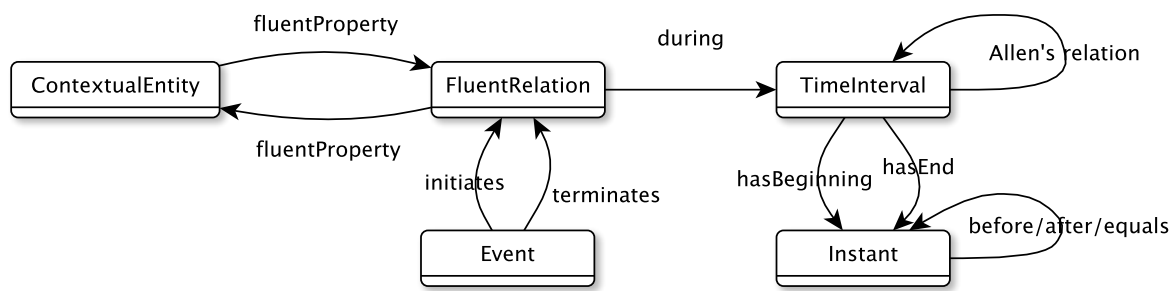

FIGURE 1 – Diagramme de classes RDFS de la représentation des fluents

Pour représenter une assertion de fluent nous utilisons un patron de relations *n-aires* (Noy et Rector, 2006) en introduisant un objet de type *FluentRelation* inspiré de (Preventis et al, 2012)), comme décrit sur la figure 1.On lui attribue le prédicat *during* et l'objet *TimeInterval* pour modéliser l'intervalle de temps durant lequel le fluent est valide. Un évènement peut initier ou terminer la période de validité (propriétés *initiates* et *terminates*). L'expression de l'intervalle de validité peut être quantitative, si l'on précise les instants de début et de fin, ou qualitative si l'on spécifie l'intervalle par ses relations de Allen (during, overlaps, meets, ...) avec d'autres intervalles.

La figure 2 montre la représentation de l'énoncé « Saussure a vécu en Allemagne entre 1876 et 1881 et en France entre 1881 et 1891 », qui contient deux assertions de fluents. Les entités contextuelles représentées dans ce schéma sont Saussure (une instance de la classe Personne) et Allemagne et France (deux instances de la classe Lieu). L'évènement qui a déclenché le premier fluent est le début des études universitaires de Saussure. Et l'évènement qui le pousse à changer son lieu de résidence est son enseignement et ses études à l'EPHE.

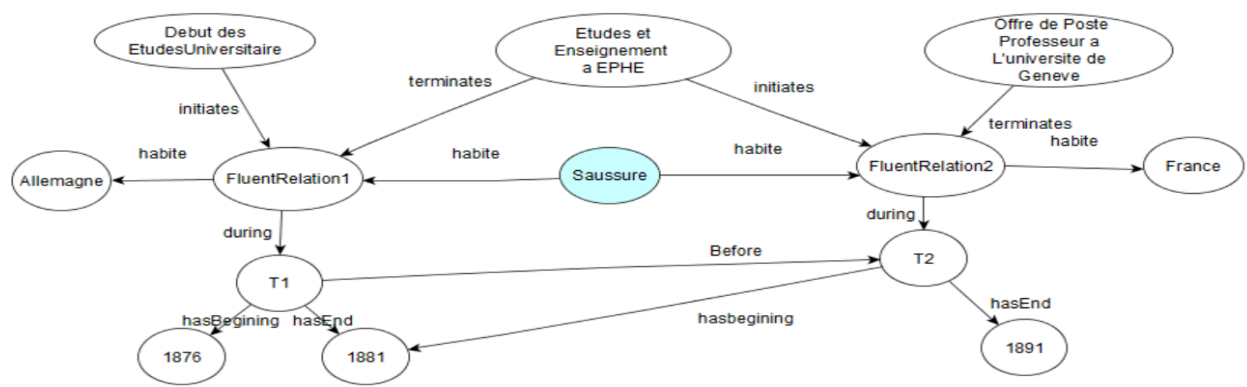

FIGURE 2 – Représentation RDF de deux assertions de fluents. Les types des objets sont omis

### 4.2 Modélisation de l'évolution terminologique

L'un des besoins importants des historiens (des sciences) concerne la représentation de la terminologique utilisée dans chaque document. En effet, dans le cas d'un auteur scientifique, les changements terminologiques sont fréquents car celui-ci travaille généralement sur des concepts encore instables ou crée lui-même de nouveaux concepts. C'est pourquoi, outre les entités contextuelles de type personne, évènement, lieu, relation scientifique, ... le modèle permet de représenter les terminologies utilisées par le ou les auteurs des documents. Chaque terminologie est formée d'entités de type concept associée à des termes et définitions.

La propriété *uses* permet de créer des assertions de fluents pour représenter le fait qu'un auteur utilise une terminologie particulière pendant une période (il n'est pas exclu qu'il en utilise plusieurs en parallèle !)

## 5 Raisonnement temporel sur le modèle

Dans cette section nous montrons comment le raisonnement temporel peut être réalisé pratiquement sur le modèle avec SWRL et SPARQL.

### 5.3 Inférences Temporelles avec SWRL et SPARQL

Il est possible d'utiliser directement des règles SWRL pour inférer des faits qui découlent des connaissances temporelles (fluents). On peut par exemple définir une règle temporelle

*Représentation des connaissances temporelles dans les documents historiques*

pour chaque propriété fonctionnelle: pour une propriété (fluente) fonctionnelle `prop` on n'a qu'une valeur pour un objet à un instant donné. Ce qui peut se traduire par la règle SWRL

```
prop(?x, ?f1, ?y1)[?i1], prop(?x, ?f2, ?y2)[?i2], overlaps(?i1, ?i2)
-> sameAs(?y1,?y2)
```

où `P(?X, ?F, ?Y)[?I]` est une abréviation de

```
FluentRelation(?F), P(?X,?F), P(?F,?Y), during(?F,?I)
```

Toutefois, les règles SWRL ne permettent pas la génération de nouveaux objets (noeuds RDF). Elle ne peuvent donc pas servir à créer de nouvelles assertions de fluents qui nécessitent la création d'instances des classes *FluentRelation* et *TimeInterval*.. Par contre, on peut utiliser des opérations SPARQL INSERT pour créer ces objets sous forme de nœuds blancs. Un exemple typique est la règle

*Si un manuscrit* M *écrit par* A *est une lettre à* B *et le temps d'écriture de* M *est* [$t_1$ .. $t_2$] *alors* A *connaît* B *pendant l'intervalle* [$t_1$ .. fin de la période considérée]

que l'on pourrait écrire en "SWRL étendu" sous la forme

```
Lettre(?l), auteur(?l, ?a), destinataire(?l, ?b), dateEcriture(?l, ?t1)
-> connait(?a, ?f, ?b)[?i], start(?i, ?t1), stop(?i, fin_periode)
```

et qui se traduit directement en SPARQL par

```
INSERT {?A :connait
         [a :FluentRelation ;
          :during [a :TimeInterval ; :start ?t1; :stop :fin_periode];
          :connait ?B]}
WHERE {?L a :Lettre. ?L :auteur ?A. ?L :destinataire ?B. ?L :
     dateEcriture ?t1}
```

On peut alors obtenir un système d'inférence complet en itérant l'exécution des règles SWRL et des insertions SPARQL de manière exhaustive.
Pour garantir que le processus se termine il faut cependant ajouter deux conditions
1. ne pas générer de nouveaux fluents superflus (qui ont les mêmes propriétés qu'un fluent déjà existant). Cette condition peut être incorporée directement dans les expressions d'insertion SPARQL sous forme d'un filtre.
2. les règles d'inférence (en SWRL étendu) ne doivent pas faire référence à des individus de *FluentRelation* ou de *TimeInterval* en position de sujet ou d'objet d'une assertion de fluent.

Ces conditions sont suffisantes car on ne considère qu'un ensemble fini d'instants (temps discret) et aucun autre type d'objet que des *FluentRelations* et *TimeInterval* n'est créé.

## 5.4 Construction de snapshots[1] pour le raisonnement synchronique

Afin de permettre le raisonnement synchronique nous avons introduit la notion de snapshot. Un *snapshot* sur un intervalle temporel [$t_1$, $t_2$] représente les faits qui restent vrais entre $t_1$ et $t_2$, c'est-à-dire tous les faits statiques et tous les fluents dont l'intervalle de validité contient [$t_1$, $t_2$]. Le snapshot est un graphe non temporel qu'on obtient en remplaçant les fluents valides sur cet intervalle par des triples non temporels, selon la règle

---

[1] Nous utiliserons le terme anglais plutôt que le terme *instantané* qui est d'usage peu courant.

```
prop(?x, ?f, ?y)[i], contains(i,[t1,t2])  -> prop(?x,?y)
```

qui s'implémente facilement par une opération SPARQL DELETE/INSERT.

Il devient alors possible d'interroger "synchroniquement" le snapshot avec des requêtes SPARQL ne faisant pas intervenir le temps ou d'appliquer des règles SWRL non temporelles.

Les faits inférés dans un snapshot sur [$t_1$, $t_2$] peuvent ensuite être réinjectés dans la base de connaissance globale sous forme de fluents valides sur [$t_1$, $t_2$]. .

## 6  Conclusion

Les ressources du domaine historique constituent un objet d'étude intéressant pour les communautés du web sémantique. Nous avons proposé dans ce papier un modèle temporel historique et son implémentation avec les techniques du web sémantique. Ce modèle met l'accent sur la cause des changements au cours du temps et sur l'inférence de nouvelles connaissances temporelles. La partie structurelle du modèle a été testé sur de nombreux cas représentatifs fournis par les experts saussuriens. Il reste nécessaire de tester des techniques de raisonnement temporel, qui nécessiteront probablement des travaux d'optimisation. D'autre part nous devrons travailler avec les utilisateurs au recensement et à l'écriture des règles d'inférence temporelle les plus pertinentes dans le domaine étudié, ce qui permettra d'évaluer l'utilisabilité du modèle pour des humanistes.